# Mini-DDSM: Mammography-based Automatic Age Estimation

CHARITHA DISSANAYAKE LEKAMLAGE

Blekinge Institute of Technology, SE-371 79 Karlskrona, Sweden

FABIA AFZAL

Blekinge Institute of Technology, SE-371 79 Karlskrona, Sweden

ERIK WESTERBERG

Blekinge Institute of Technology, SE-371 79 Karlskrona, Sweden

ABBAS CHEDDAD*

Blekinge Institute of Technology, SE-371 79 Karlskrona, Sweden

Age estimation has attracted attention for its various medical applications. There are many studies on human age estimation from biomedical images. However, there is no research done on mammograms for age estimation, as far as we know. The purpose of this study is to devise an AI-based model for estimating age from mammogram images. Due to lack of public mammography data sets that have the age attribute, we resort to using a web crawler to download thumbnail mammographic images and their age fields from the public data set; the Digital Database for Screening Mammography. The original images in this data set unfortunately can only be retrieved by a software which is broken. Subsequently, we extracted deep learning features from the collected data set, by which we built a model using Random Forests regressor to estimate the age automatically. The performance assessment was measured using the mean absolute error values. The average error value out of 10 tests on random selection of samples was around 8 years. In this paper, we show the merits of this approach to fill up missing age values. We ran logistic and linear regression models on another independent data set to further validate the advantage of our proposed work. This paper also introduces the free-access Mini-DDSM data set.

---

* Corresponding author: SMIEEE, ACM, email: abbas.cheddad@bth.se.

# 1 INTRODUCTION

Biomedical imaging is a technique of creating a visual representation of the body that can be used for medical diagnoses and clinical analyses. Biomedical imaging involves the use of various technologies such as X-rays, CT-scans, magnetic resonance imaging (MRI), ultrasound, mammography, light (endoscopy, OCT) or radioactive pharmaceuticals (nuclear medicine: SPECT, PET) for diagnosing and helping to treat medical conditions of patients more efficiently. Along with that biomedical images can be used for estimating the age of a person. We were able to find many studies on human age estimation from face images, dental images, MRI, X-rays etc for different purposes. For example, in one of the studies [1], researchers work on predicting the age of a patient from a chest X-ray by using deep learning methods. Similarly, some researchers presented a software-based solution for estimating age automatically based on 3D MRI images of the hand [2]. Bone age assessment (BAA) can be useful in a variety of situations. For example, it can be used to predict how much longer children will grow when they will hit puberty or even their final height [3]. It can also be used to monitor the progress of children being treated for conditions that affect growth. BAA is also very useful when it comes to identifying people lacking proper identification [4]. In recent years, there has been a significant increase in the number of refugees lacking proper identification seeking asylum in Europe. Unaccompanied individuals under the age of 18 are eligible for special rights according to the United Nations Convention on Rights of the Child, so from a legal standpoint, an accurate assessment is important to create a fair process.

There are two methods used for manual BAA: The Greulich-Pyle (GP) and the Tanner-Whitehouse methods (TW) [5]. In 2009, Thodberg et al. introduced a fully automated method for determining skeletal maturity: BoneXpert [6]. It used methods from statistics and machine learning (ML), which at the time had not been used for this task. The BoneXpert system divides the BAA into three steps. The first step reconstructs the bone borders, which are used for the assessment. The second step computes the bone age values for each area. The third step converts the bone age values into either GP or TW-bone age using simple postprocessing.

This software is licensed on a pay-per-analysis basis. The BoneXpert connects as a DICOM node to a DICOM network. To perform a BAA, a hand radiograph is pushed from the PACS (image Archiving and Communication System) to the BoneXpert DICOM node. The BoneXpert server performs the assessment and then returns an annotated image to PACS and stores it next to the original radiograph. The BoneXpert is not an open source, meaning that we do not know what is going on internally during the BAA. It is also relatively expensive. In 2011, the price was 10 euros per assessment [7]. Current pricing is to our knowledge not available through their website.

In 2017, a competition was hosted by Radiological Society of North America (RSNA) in which the competitors set out to determine the bone age of a person based on their hand radiograph [8] using machine learning techniques. The competitors used a data set provided through KAGGLE. The competition was judged based on the competitors best provided mean absolute distance (MAD). A visualization of an example of such frameworks is shown in Figure 1.

There is a decent number of studies related to age estimation from biomedical images. However, to our knowledge there is no research done related to age estimation based on mammographic images, be it tomosynthesis, MRI or X-ray based (digitized films or full field digital mammograms -FFDM).

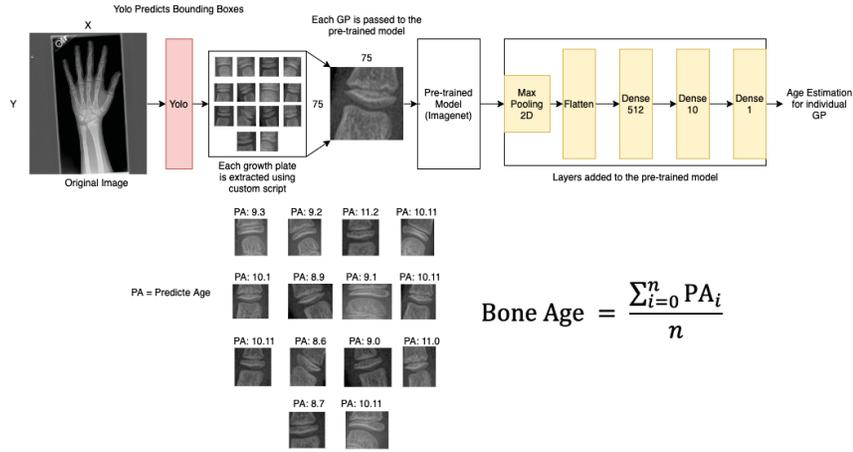

Figure 1: An example of a complete workflow of age estimation through hand's growth plates [9]

To our knowledge, all the previous work pertaining to mammography analysis focused on segmentation and/or classification tasks. Therefore, in this paper, we focus on mammogram X-ray images to perform nonlinear regression in the aim to automatically estimate age of a woman. Age is an important factor in risk prediction of breast cancer.

### 1.1 Motivation

From the existing literature we reviewed, we found that there are several studies related to age estimation based on biomedical imaging such as X-ray images of hand, teeth, knees etc. or studies related to Magnetic resonance imaging (MRI) of the brain and facial images. However, we were unable to find any research based on mammogram images in conjunction with automatic age estimation, thus a research gap that we wanted to fill.

We anticipate that this research provides a useful tool in filling out missing age values (*null* values) in some mammography-based cohorts. These *null* values can be a result of different reasons. For example, it is possible that mistakes may occur during data collection/entry, or in underdeveloped countries existing mammography cohorts might have missing age attribute entries due perhaps to patient's unwillingness to reveal her real age, or because of a system glitch/file corruption, or because of relying on a traditional complicated bureaucratic system to extract paper-based birth certificates to prove age, etc. In such scenarios, it is common for researchers to either discard the entire row consisting of these *null* values or to take the average mean value of age to substitute the *null* age entries. Both resolutions are not optimal to fill up missing data. This will result in a waste of resources as well as affecting the prediction power in many studies which consider age as a factor. For example, in breast cancer studies, it is proven that the age of a patient is a risk factor of developing breast cancer [10], [11], [12], [13]. Therefore, it is crucial for such studies to have age values as accurate as possible rather than resorting to use the average value.

## 1.2 Contribution

In a nutshell, the contribution of this work is twofold:

Exploring the possibility to estimate women's age from their acquired mammographic images, which we believe is a new explored modality worth further investigation.

Providing a free and easy access to the known Digital Database for Screening Mammography (DDSM) which we termed Mini-DDSM (i.e., thumbnails version of DDSM) with the age attribute and classification (benign, malignant and normal). The full data set is available for free download at: https://ardisdataset.github.io/MiniDDSM/

## 2 METHODOLOGY

Before delving into the methods, we opt to first justify the need for creating the Mini-DDSM.

### 2.1 Why Mini-DDSM?

First, we should stress that there is a real scarcity of publicly free mammographic data sets. The data sets that have been extensively used by the research community are the Mammographic Image Analysis Society (MIAS) [14] database, the Digital Database for Screening Mammography (DDSM) [15] and the Digital Database for Screening Mammography (DDSM) called CBIS-DDSM [16]. Among these data sets, only the DDSM contains the age attribute associated with each mammographic image.

Second, and to answer the nagging question why Mini-DDSM, it is important to know that the DDSM database has a website maintained at the University of South Florida for purposes of keeping it accessible on the web. However, image files are compressed with lossless JPEG (i.e., ".LJPEG") encoding that are generated using a broken software (or at least an outdated tool as described on the DDSM website). CBIS-DDSM provides an alternative host of the original DDSM, but unfortunately, images are stripped from their original identification filename and from the age attribute.

Hence, the only option we had, though not optimum, was to create a web crawler script that downloaded all thumbnail images from the DDSM website along with the filename (i.e., file ID and whether it is a L/R CC or L/R MLO image), the status (i.e., benign, malignant, normal), and most important to this study, the patient's age. This Mini-DDSM was manually examined to remove any outliers (e.g., a 1-year old woman with a mammogram). The data set we collected is described as shown in Table 1 and Figure 2. Image dimension is a variable between 125 and 320 pixels.

Table 1: Mini-DDSM Data set Characteristics.

| Label  | #images | μ (age) | σ (age) |
|--------|---------|---------|---------|
| Normal | 2728    | 57.8284 | 11.6651 |
| Cancer | 3596    | 61.3971 | 12.7968 |
| Benign | 3360    | 53.1036 | 12.0260 |
| Total  | 9684    | 57.5143 | 12.7149 |

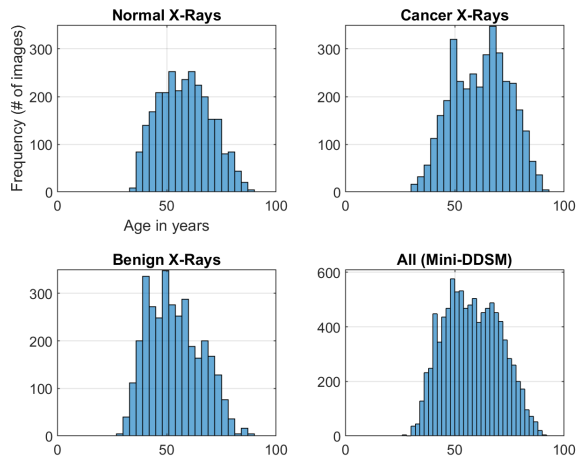

Figure 2: Age distribution in the Mini-DDSM data set

### 2.2 Age Estimation using ResNet Deep Features

ResNet was introduced in 2015 and won the first place in the ILSVRC competition on tasks of ImageNet localization, ImageNet detection, COCO segmentation and COCO detection [17]. ResNet embodies a deep architecture which is designed on the principle of residual learning. There are several versions that belong to the family of ResNet such as ResNet-101, ResNet-50, ResNet-34, ResNet-18 and ResNet-152. The number of layers in each architecture dictates the unique name of the network [18]. Othmani et al. [19], presented a comparative analysis of several deep learning architectures for automatic age estimation (AAE) from facial images in which ResNet50 was amongst the best performers. Sabottke et al. [20], used ResNet50 for AAE based on chest radiography. They were able to achieve a mean absolute error (MAE), equivalent to MAD, of around 4.94 years.

Both studies are published in 2020, which motivated us to use ResNet50 as the chosen deep learning model for feature extraction. We use transfer learning by loading the weights of the ResNet50 pre-trained network. The input images were resized to (224x224x3) to match the input layer size by duplicating the gray image to form the RGB tuples.

### 2.3 AAE Regression Model

The selected regression approach is the Random Forests (RF) [21] non-linear regressor, which can help reduce the effects of overfitting and improve generalization. RF is also known to scale up well with large data sets, while it maintains a good performance when having limited training sample size.

First, we randomly selected an equal number of images from each category (Normal, Cancer, Benign) to eradicate any sample size bias. Thus, the sample size was 2728 from each folder which totalled up to 8184. Subsequently, we run tests for 10 times, each of which used a different data set by randomly dividing the samples into 70% for training and 30% for testing while controlling the random number generator seed for reproducibility. The cancer status was not a factor to include in the regression equation since we want to devise a generic model for AAE not for classification of disease status.

## 3 EXPERIMENTS AND RESULTS

This section serves two purposes. First, we show the performance of our developed RF model in AAE. Second, we take this experiment one step further by estimating the age for women in the renown MIAS database (which does not have age attribute) and we construct logistic and linear regression models to pinpoint the benefits of our proposed approach. The experiment A was carried out in "MATLAB" (ver. R2019b) and experiments B&C were conducted in the statistical computing environment "R".

### 3.1 RF Automatic Age Estimation on the Mini-DDSM

The average MAE value for the RF model of 10 random tests is around ±8 (years). The obtained results for a single run are depicted in Figure 3.

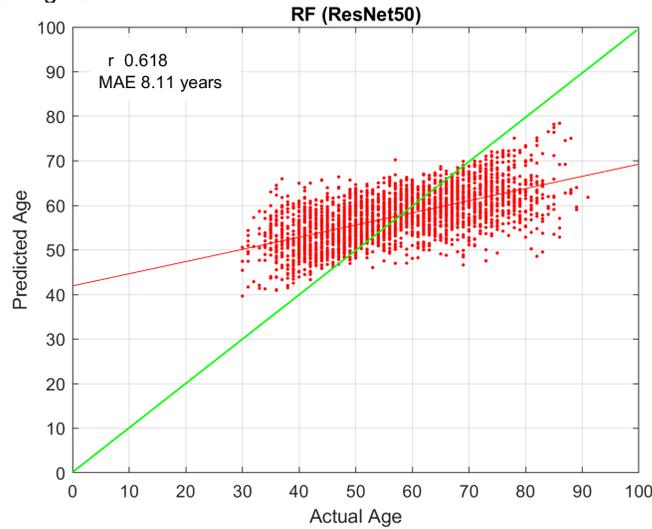

Figure 3: Actual age as opposed to the predicted age. The value r denotes the Pearson's linear correlation coefficient

This implies that a baseline approach of using the average age of all samples as the estimated age would result in a MAE of around 10.6 years, which makes the MAE of our ML-based approach closer to the actual age by 2.6 years on average. This is an apparent improvement which can be further boosted via several suggested considerations as we will highlight in the discussion section (Sec. 4).

### 3.2 Logistic Regression - MIAS Database-

We examined the association between the predicted age for MIAS database along with measures that came with the database and breast cancer status (case–control analysis), based on the data set of 51 cases (malignant) and 271 controls (we aggregated both benign and normal samples to increase sample size), using unconditional logistic regression (case–control status as the dependent variable and age as the independent variable, adjusting for percent density (PD), fibroglandular tissue presence (FGT), breast size (BS), mean intensity of the breast (MIB) and mean intensity of pectoral muscle (MIP) [22]). Effect estimates are presented as odds ratio (ORs) in Table 2. The FGT (2[nd] column in the MIAS data set), namely, Fatty (F), Fatty-glandular (G) and Dense-glandular (D) were coded 0/1/2, treated as a continuous variable, respectively.

Table 2: Effect estimates for different mammographic measurements on case–control status, n = 322 (51 cases and 271 controls). Point estimates, interval estimates and p-values (Wald tests) are based on estimated coefficients for the measurements in logistic regression models with case-control status as outcome.

| Covariate | Estimate (95% CI) | *p*-value |
|---|---|---|
| Age_predicted | -0.0769 (-0.1794, 0.0256) | 0.134 |
| PD | 0.0539 (-2.8121, 2.9199) | 0.970 |
| FGT | -0.3284 (-0.8432, 0.1864) | 0.202 |
| BS | 3.715e-07 (-2.6×$10^{-6}$, 3.4×$10^{-6}$) | 0.805 |
| MIB | 4.4410 (-7.5350, 16.4170) | 0.458 |
| MIP | 0.9398 (-5.1562, 7.0358) | 0.758 |

It is evident that none of the covariates reveal a strong statistical significance, this is partially due to the small number of samples in the MIAS data set and the benign cases are considered as cancer cases. Nevertheless, as stated earlier, our aim here is not to devise a breast cancer classifier, rather we are after associating a value to our developed age estimation model. Since MIAS has no age values associated with its mammograms, we used our developed model to estimate the age and simulate the missing age entries.

To this end, we conducted an extra experiment. In this experiment, we calculated the average of age column and assigned that randomly to several age entries to check its impact on the statistical significance. Table 3 summarized the outcome, corresponding to the set-up in Table 2 but for conciseness, we only display the Age covariate. Aside from the *p-value*, we also infer the *Akaike information criterion* (AIC). When comparing fitted models, the smaller the AIC, the better is the fit. The AIC is a surprisingly simple estimator of the average out-of-sample prediction error [23].

Table 3: Effect estimates for the age average measurement on case–control status, n = 322 (51 cases and 271 controls). Age_xx, where xx denotes the number of random age entries replaced with average age value.

| Covariate | AIC | *p*-value |
|---|---|---|
| Age_predicted | 289.89 | 0.134 |
| Age_36 | 291.2 | 0.320 |
| Age_63 | 291.32 | 0.352 |
| Age_80 | 291.44 | 0.386 |

### 3.3 Linear Regression - MIAS Database-

We fitted linear regression models using the age and age average measures one at a time as the independent variable and FGT as the outcome variable and carried out Wald tests. Testing this association, $\varphi(age, FGT)$ seems natural since physiologically speaking, the fibroglandular tissue decreases by age as a result of fat invasion. Table 4 shows the results. As can be seen from the table, the more added average age values, the weaker the p-value will be. This proves that the estimated AAE using our approach, though not perfect due to lack of training data, seems to outperform the average mean approach [24].

Table 4: p-values assessing the association of the mammography automatic age estimation with the FGT.

| Covariate | AIC | Estimate (95% CI) | *p*-value |
|---|---|---|---|
| Age_predicted | 748.21 | -0.0894 (-0.1150, -0.0638) | 1×$10^{-11}$ |
| Age_63 | 760.37 | -0.0848 (-0.1134, -0.0567) | 7×$10^{-9}$ |
| Age_80 | 759.97 | -0.0896 (-0.1196, -0.0596) | 6×$10^{-9}$ |
| Age_112 | 769.83 | -0.0800 (-0.1121, -0.0480) | 9×$10^{-7}$ |

## 4 DISCUSSION

The experiments that we carried out with the developed mammography based AAE reveal the potential of our proposed approach. The impact of which would have been even greater if we have had access to adequate data sets (i.e., high resolution mammographic images with age attribute). We relied on publicly available data sets despite their scarcity. We envision that the proposed approach would benefit from the following:

### 4.1 High Resolution Mammograms

The AAE model that we derived by training ResNet50 deep features extracted from DDSM thumbnails was useful to establish a proof-of-concept, however, if our approach needs to be deployed in real-life, access to high resolution mammograms along with error-free age attribute is deemed vital. This will, with overwhelming probability, immensely increase the correlation of the actual age and the predicted age shown previously in Figure 3.

### 4.2 Large Case-control Cohort

This is an expected requirement to reach to a statistical significance. The MIAS data set, unfortunately, provides only 51 detected cancer cases (malignant) which are insufficient to conduct a large-scale data analysis. Thus, our approach needs validation on external larger data sets. As we speak, the MIAS and the mini-DDSM (thumbnails) are the only options we have.

### 4.3 Beyond AAE

Our approach is viable not only for age estimation, but the approach can be adapted (retraining needed) to estimate body mass index (BMI) for example, which is normally collected from patients when undertaking their first mammography screening.

## 5 CONCLUSION

In this paper, we propose an automatic age estimation (AAE) via patient's mammographic image. The ResNet50-based regression model was trained on the Mini-DDSM (thumbnail images retrieved using web crawling from the DDSM webpage). The Mini-DDSM data set as well as the associated and data are shared freely via the following link: https://ardisdataset.github.io/MiniDDSM/ .

There exist techniques to estimate age from hand [25], wrist [26], knee [27], [28] and clavicle [29], however, to the best of our knowledge, our contribution of AAE from mammograms is new. Our study has the potential to serve the research community in the field of breast cancer research where age column, in a given data set, contains missing age values (*null* values). Our approach bridges the age gap (entries with missing values) in a more sensible way than by bridging that with the overall average age value or by discarding these entries which are common practices when encountered with data having some missing values.

## ETHICS STATEMENT

This study is built on a publicly available data set not linked to individuals (e.g., Personal data had been de-identified or anonymized). According to the definition of the GDPR (the European General Data Protection Regulation), the data set is not considered as a personal information belonging to any individual. Therefore, there shall be no ethical issues.

**ACKNOWLEDGMENTS**

This work is partially funded by the research project scalable resource efficient systems for big data analytics by the Knowledge Foundation (Grant: 20140032) in Sweden.